\documentclass[a4paper,5p,fleqn]{cas-dc}
\usepackage[numbers]{natbib}
\usepackage{multirow}
\usepackage[table]{xcolor}
\definecolor{lightgreen}{RGB}{144, 238, 144}
\definecolor{lightorange}{RGB}{255, 200, 100}
\definecolor{lightred}{RGB}{255, 100, 100}

\def\tsc#1{\csdef{#1}{\textsc{\lowercase{#1}}\xspace}}
\tsc{WGM}
\tsc{QE}
\tsc{EP}
\tsc{PMS}
\tsc{BEC}
\tsc{DE}

\begin{document}
\let\WriteBookmarks\relax
\def\floatpagepagefraction{1}
\def\textpagefraction{.001}
\shorttitle{Cystic Hygroma Detection Using USF-MAE}
\shortauthors{Y. Megahed et~al.}

\title [mode = title]{Improved cystic hygroma detection from prenatal imaging using ultrasound-specific self-supervised representation learning}                      



\author[1,2]{Youssef Megahed}[type=editor,
                        orcid=0009-0004-2595-5468]
                        
\cormark[1]
\ead{youssefmegahed@cmail.carleton.ca}

\credit{Conceptualization of this study, Methodology, Software}

\affiliation[1]{organization={Department of Systems and Computer Engineering, Carleton University}, 
                city={Ottawa},
                state={Ontario},
                country={Canada}}
                
\affiliation[2]{organization={Department of Methodological and Implementation Research, Ottawa Hospital Research Institute}, 
                city={Ottawa},
                state={Ontario},
                country={Canada}}

\affiliation[3]{organization={Department of Acute Care Research, Ottawa Hospital Research Institute}, 
                city={Ottawa},
                state={Ontario},
                country={Canada}}

\affiliation[4]{organization={Children's Hospital of Eastern Ontario Research Institute}, 
                city={Ottawa},
                state={Ontario},
                country={Canada}}

\affiliation[5]{organization={Better Outcomes Registry \& Network Ontario, Children’s Hospital of Eastern}, 
                city={Ottawa},
                state={Ontario},
                country={Canada}}

\affiliation[6]{organization={Department of Obstetrics and Gynecology, University of Ottawa}, 
                city={Ottawa},
                state={Ontario},
                country={Canada}}

\affiliation[7]{organization={School of Epidemiology and Public Health, University of Ottawa}, 
                city={Ottawa},
                state={Ontario},
                country={Canada}}
                
\affiliation[8]{organization={Department of Obstetrics, Gynecology \& Newborn Care, The Ottawa Hospital}, 
                city={Ottawa},
                state={Ontario},
                country={Canada}}

\affiliation[9]{organization={International and Global Health Office, University of Ottawa}, 
                city={Ottawa},
                state={Ontario},
                country={Canada}}

\affiliation[10]{organization={Department of Clinical Science and Translational Medicine, University of Ottawa}, 
                city={Ottawa},
                state={Ontario},
                country={Canada}}

\author[3]{Robin Ducharme}
\author[3]{Inok Lee}
\author[6,8]{Inbal Willner}
\author[1]{Adrian D. C. Chan}
\author[3,4,5,6,7,8,9]{Mark Walker}
\author[1,2,4,7,10]{Steven Hawken}
\cormark[2]

\ead{shawken@ohri.ca}


\credit{Data curation, Writing - Original draft preparation}


\cortext[cor1]{Corresponding author}
\cortext[cor2]{Principal corresponding author}

\begin{abstract}
Cystic hygroma is a high-risk prenatal ultrasound finding that portends high rates of chromosomal abnormalities, structural malformations, and adverse pregnancy outcomes. Automated detection can increase reproducibility and support scalable early screening programs, but supervised deep learning methods are limited by small labelled datasets. This study assesses whether ultrasound-specific self-supervised pretraining can facilitate accurate, robust deep learning detection of cystic hygroma in first-trimester ultrasound images.

We fine-tuned the Ultrasound Self-Supervised Foundation Model with Masked Autoencoding (USF-MAE), pretrained on over 370,000 unlabelled ultrasound images, for binary classification of normal controls and cystic hygroma cases used in this study. Performance was evaluated on the same curated ultrasound dataset, preprocessing pipeline, and 4-fold cross-validation protocol as for the DenseNet-169 baseline, using accuracy, sensitivity, specificity, and the area under the receiver operating characteristic curve (ROC-AUC). Model interpretability was analyzed qualitatively using Score-CAM visualizations.

USF-MAE outperformed the DenseNet-169 baseline on all evaluation metrics. The proposed model yielded a mean accuracy of $0.96 \pm 0.02$, sensitivity of $0.94 \pm 0.06$, specificity of $0.98 \pm 0.02$, and ROC-AUC of $0.98 \pm 0.02$ compared to $0.93 \pm 0.03$, $0.92 \pm 0.07$, $0.94 \pm 0.01$, and $0.94 \pm 0.03$ for the DenseNet-169 baseline, respectively. Qualitative Score-CAM visualizations of model predictions demonstrated clinical relevance by highlighting expected regions in the fetal neck for both positive and negative cases. Paired statistical analysis using a Wilcoxon signed-rank test confirmed that performance improvements achieved by USF-MAE were statistically significant ($p = 0.0057$).

Ultrasound-specific self-supervised pretraining enables deep learning models to detect cystic hygroma with higher accuracy and robustness than supervised convolutional neural networks trained from scratch. This data-efficient approach requires minimal manual annotation and demonstrates the potential of self-supervised learning for scalable decision support in early prenatal anomaly screening.
\end{abstract}





\maketitle

\section{Introduction}
Ultrasound is a commonly used screening and diagnostic tool for prenatal conditions. It is a real-time and radiation-free imaging modality, yet fetal ultrasound images are difficult to interpret due to high noise, dependency on the operator, and small fields of view {\cite{b1}}. The quality of the ultrasound scan highly depends on the sonographer’s experience and the fetal position. This makes it difficult to acquire high-quality images, and results in high inter-observer variability {\cite{b1,b2,b33,b11,b12,b13}}. In addition, sonographic images are typically of low contrast and are speckled. As a result, sonographers often have difficulties recognizing or identifying abnormalities in the images {\cite{b3}}. This can lead to an increased risk of false-negative or false-positive findings for clinically significant abnormalities, creating a large need for assistive technology in fetal anomaly detection.

In recent years, artificial intelligence (AI) techniques, and more specifically deep learning (DL), have shown promise in improving obstetric ultrasonography {\cite{b2,b33,b8,b14}}. DL models are well-suited for pattern recognition and have been used to classify ultrasound images (e.g., normal vs. abnormal) and to localize or segment fetal structures {\cite{b2}}. In prenatal imaging, many studies have shown the feasibility of supervised DL for detecting various fetal anomalies, including cardiac, central nervous system, renal, and respiratory anomalies, among others {\cite{b3}}. For example, fetal brain structures were identified on standard plane scans by Lin \textit{et al.} {\cite{b16}} using a convolutional network. A more recent system by Xie \textit{et al.} {\cite{b15}} also detected multiple intracranial anomalies using YOLOv3, a state-of-the-art object detector, and provided simultaneous localization for ten different anomalies. These works, among others, point to AI's potential role in supporting sonographers' decision-making and reducing the subjectivity of prenatal anomaly detection. However, AI for obstetric ultrasound is still in its early stages, and further work is needed to more rigorously validate AI models for generalizability and clinical use {\cite{b2}}.

\begin{figure*}
	\centering
	\includegraphics[width=.98\textwidth]{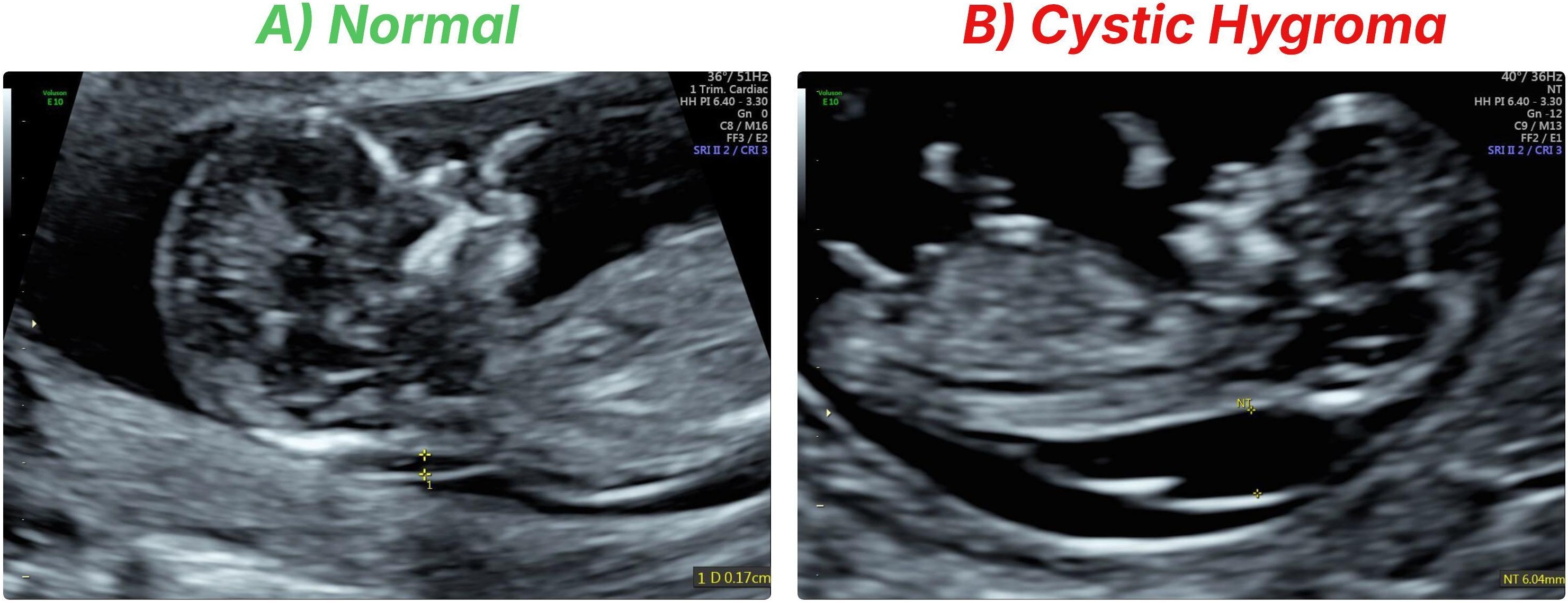}
	\caption{Comparison of first-trimester fetal ultrasound images. A) Normal fetus with typical NT thickness. B) Fetus with cystic hygroma, showing a markedly enlarged, multiloculated nuchal fluid collection consistent with lymphatic malformation.}
	\label{FIG:class_examples}
\end{figure*}

One important prenatal finding on obstetric sonography is cystic hygroma. Cystic hygroma is a lymphatic lesion, appearing as a fluid-filled, multiseptated cystic mass, that is often seen on the fetal neck region, also known as the nuchal area. It is an important sonographic finding due to its strong correlation to severe fetal pathology. Cystic hygroma in the first trimester is highly predictive for aneuploidies (or abnormal number of chromosomes). In fact, over 50\% of fetuses with first-trimester cystic hygroma will have aneuploidy, including Down syndrome (trisomy 21), Turner syndrome (45, X), or trisomy 18 {\cite{b4}}. Cystic hygroma also has a high co-occurrence with other structural malformations (e.g., cardiac defects) and is highly predictive for fetal demise {\cite{b4}}. In fact, cystic hygroma is one of the highest prenatal risks for genetic syndromes and poor outcomes when compared to a simple increased Nuchal Translucency (NT) {\cite{b4}}. As a result, early identification of cystic hygroma on ultrasound can play an important role in prenatal prognostication and perinatal management. In practice, cystic hygroma is usually visualized as an enlarged NT ({Fig~\ref{FIG:class_examples}B}) with septations on the 11-14 week scan, and is often thicker than the normal cut-off for gestational age {\cite{b3}}. The prevalence of cystic hygroma has been reported to be around 1 in 285 pregnancies in a general obstetric population {\cite{b3}}, so it is an uncommon, but important, sonographic finding to be able to identify. This prenatal finding often results in invasive testing, genetic counselling, and may even impact management of the pregnancy, highlighting the importance of accurately and consistently detecting cystic hygroma.

Recently, there have been efforts to apply supervised DL to automate cystic hygroma diagnosis from ultrasound images. Walker \textit{et al.} {\cite{b5}} developed a convolutional neural network (CNN) based on the DenseNet-169 architecture to distinguish first-trimester mid-sagittal scans as either normal or showing a cystic hygroma (an example of each class is shown in {Fig~\ref{FIG:class_examples}}) {\cite{b5}}. Their model was trained on a curated dataset of 289 ultrasound images (129 cystic hygroma cases, 160 normal controls) and achieved 93\% classification accuracy (sensitivity 92\%, specificity 94\%) in distinguishing cystic hygromas from normal fetuses {\cite{b5}}. This proof-of-concept study demonstrated that DL can attain expert-level performance in early anomaly detection. Moreover, Gradient-CAM {\cite{b17}} visualizations confirmed that the model's decisions were driven by the fetal neck region, aligning with clinical expectations {\cite{b5}}. While encouraging, the DenseNet-based approach has notable limitations. One of them is that the DenseNet-169 model had to be trained from scratch on ultrasound images because conventional transfer learning from natural image data (ImageNet) {\cite{b18}} was deemed suboptimal due to the domain mismatch {\cite{b5}}. Ultrasound images have fundamentally different textures and content compared to everyday photographs, so an ImageNet-pretrained CNN may not provide useful features {\cite{b5}}. Training a deep CNN without any pretrained initialization on only a few hundred images is far from ideal – it risks converging to local minima or learning spurious features. In Walker \textit{et al.}'s study {\cite{b5}}, extensive data augmentation and cross-validation were employed to mitigate this risk {\cite{b5}}, but the situation underscores a general challenge: supervised DL in medical imaging is often bottlenecked by scarce labelled data and poor transferability of models pretrained on non-medical images {\cite{b1}}. These limitations motivate the exploration of approaches that can leverage the abundance of unlabeled medical images to learn robust representations, reducing the dependency on large labelled datasets.

Self-supervised learning (SSL) has emerged as a powerful paradigm to address the data scarcity problem in medical imaging. In SSL, models learn from unlabeled data through surrogate objectives (pretext tasks), such as predicting missing parts of the input or distinguishing augmented views of the same image, thereby encoding meaningful features that can be fine-tuned for downstream diagnosis. A variety of SSL frameworks have shown remarkable success in computer vision in recent years. For instance, contrastive methods like SimCLR {\cite{b6}}, Momentum Contrast (MoCo) {\cite{b7}}, and Bootstrap Your Own Latent (BYOL) {\cite{b8}} train an encoder by maximizing agreement between different augmented versions of an image or by using momentum-updated networks, yielding representations that rival those learned with full supervision. SimCLR in particular demonstrated that purely self-learned features can achieve ImageNet classification accuracy on par with a supervised ResNet, given sufficient data and training time {\cite{b6}}. MoCo, on the other hand, introduced a dynamic dictionary with a queue of negative samples and a momentum-updated encoder, enabling contrastive learning with a consistent memory bank and showing excellent transfer learning performance on detection tasks {\cite{b7}}. Meanwhile, BYOL proved that negative pairs are not even required – by iteratively bootstrapping two networks (online and target) to predict each other's latent outputs, BYOL achieved state-of-the-art results without contrastive comparisons {\cite{b8}}. Beyond contrastive approaches, generative and distillation-based SSL methods have also gained traction. Masked Autoencoders (MAE) learn by masking random patches of an image and training a model to reconstruct the missing content {\cite{b9}}. This simple reconstruction task, when applied at scale, produces rich high-level representations; notably, He \textit{et al.} {\cite{b9}} reported that a ViT-Huge model pretrained with MAE reached 87.8\% ImageNet top-1 accuracy, outperforming a supervised baseline and exhibiting strong transfer to diverse tasks {\cite{b9}}. Another noteworthy strategy is the self-distillation approach exemplified by DINO {\cite{b10}}, which uses a teacher-student setup (with no labels) to train Vision Transformers (ViTs) {\cite{b26}}. DINO-pretrained ViT models not only excel in image classification (e.g., 80.1\% ImageNet top-1 with ViT-Base) {\cite{b10}}, but interestingly also yield emergent semantic segmentation capabilities from the learned attention maps {\cite{b10}}. The success of these frameworks underscores the promise of SSL for medical imaging, where unlabeled images are plentiful but expert-labelled data is limited. By leveraging SSL, one can produce a pretrained "feature extractor" model that has learned the modality-specific patterns (in our case, sonographic textures and anatomical variations) without any manual annotations, and then fine-tune this model on a small labelled dataset for the target diagnostic task {\cite{b1}}.

\begin{figure*}
	\centering
	\includegraphics[width=0.98\textwidth]{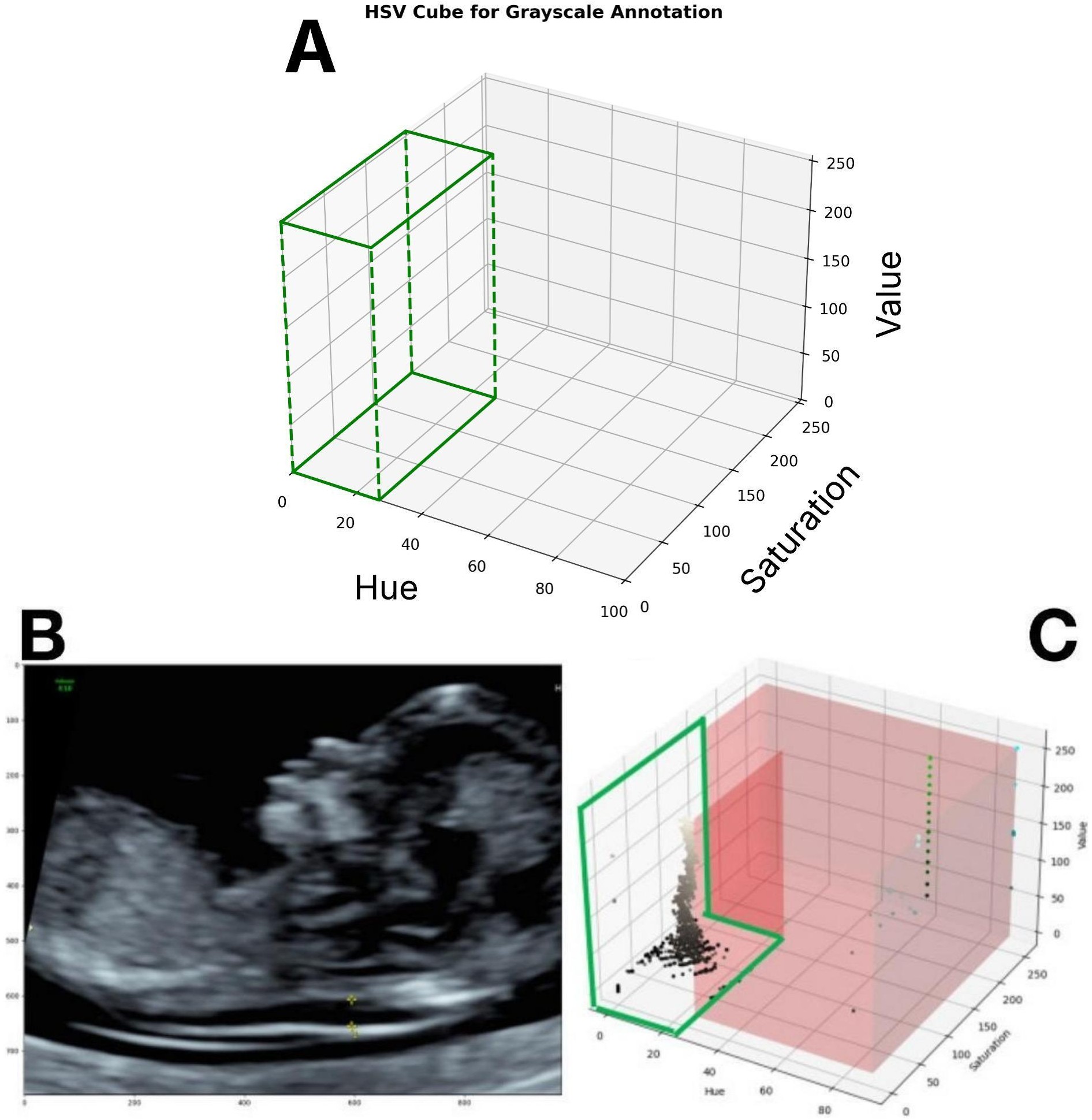}
	\caption{(A) Three-dimensional representation of the permitted grayscale area in HSV space according to the hue, saturation, and value thresholds. (B) The original ultrasound image with colored annotations. (C) The same image is seen in HSV space, with pixels outside the grayscale area highlighted to demonstrate how the preprocessing technique removes annotation artifacts. adapted from {\cite{b5}} by Walker \textit{et al.}}
	\label{FIG:preprocessing_example}
\end{figure*}

In the domain of ultrasound, this approach has recently been realized through the development of an Ultrasound Self-Supervised Foundation Model by our research group {\cite{b1}}. USF-MAE (Ultrasound Self-Supervised Foundation Model with Masked Autoencoding) was introduced as the first large-scale MAE pretraining framework devoted entirely to ultrasound imaging {\cite{b1}}. In this method, a ViT encoder {\cite{b26,b27}} was trained to reconstruct masked patches of ultrasound images, using a massive corpus of over 370,000 ultrasound frames from 46 datasets ("OpenUS-46") spanning more than 20 different anatomical regions {\cite{b1}} ({Fig~\ref{FIG:model_arc}}). By learning to fill in missing parts of hundreds of thousands of sonograms, the model distilled a broad spectrum of ultrasound-specific features that are generalizable across anatomy and pathology. The resulting USF-MAE encoder, effectively a foundation model for ultrasound, can be fine-tuned on specific downstream tasks with minimal labelled data. Megahed \textit{et al.} {\cite{b1}} report that USF-MAE, after pretraining, achieved superior results on multiple classification benchmarks (breast tumour classification, ovarian tumour classification, and gastrointestinal stromal tumor classification) compared to conventional CNN and vanilla ViT models pretrained on natural images, or even the UltraSam model {\cite{b19}}, which is a supervised learning model trained on ultrasound-specific images. For example, fine-tuning USF-MAE yielded F1-scores of 79-82\% on these tasks, outperforming standard ImageNet-pretrained ResNet and ViT or ultrasound-pretrained UltraSam benchmarks under the same conditions {\cite{b1}}. Importantly, when we applied the USF-MAE pretrained encoder to the task of fetal brain anomaly detection (ventriculomegaly), the model's performance exceeded that of models trained with generic ImageNet pretraining {\cite{b2}}. In our previous study, the USF-MAE–based classifier reached an F1-score of about 91.78\% on detecting ventriculomegaly, notably higher than the F1 achieved by a baseline ResNet-50 (around 89.22\%) or a ViT-B/16 without ultrasound-specific pretraining (79.85\%) {\cite{b2}}. The USF-MAE model also demonstrated excellent precision (94\%+) and near-perfect specificity in that application {\cite{b2}}, and its attention maps using Eigen-CAM {\cite{b20}} focused on clinically relevant regions (the fetal lateral ventricles), providing a level of explainability to its predictions. These findings illustrate the power of domain-specific self-supervised pretraining: by starting from features tailored to ultrasound, the fine-tuned model achieves higher accuracy and better generalization than was possible with limited data alone.

Building on this progress, our current work proposes to leverage the USF-MAE foundation model for the diagnosis of cystic hygroma from the same first-trimester ultrasound images used in our previous study {\cite{b5}}. In contrast to our prior approach of training a DenseNet-169 from scratch on a small dataset {\cite{b5}}, we employ USF-MAE as a pretrained initialization that already encapsulates generic fetal ultrasound feature representations. We then fine-tune this model on the same curated set of NT ultrasound images to detect cystic hygromas. The expectation is that the rich pretrained features from USF-MAE will enable more robust and data-efficient learning, thereby improving detection sensitivity and specificity for cystic hygromas even with the modest dataset available. By using a transformer-based architecture pretrained on diverse ultrasound data, we also aim to mitigate the generalization issues of earlier CNN-based methods. We hypothesize that fine-tuning a dedicated ultrasound foundation model (USF-MAE) will outperform conventional supervised CNNs in identifying cystic hygroma, leading to a more accurate and generalizable tool for early fetal anomaly screening.

\section{Methodology}
\subsection{Study Population and Image Collection}
This study represents a retrospective review of first-trimester fetal ultrasound images from the same dataset and with the same study protocol that was previously described by Walker \textit{et al.} {\cite{b5}}. All images were collected as part of routine prenatal screening exams between 11 and 14 weeks of gestation at The Ottawa Hospital, a tertiary care medical centre in Ottawa, Ontario, Canada, between March 2021 and June 2021.

\begin{table*}
\caption{\textbf{Data distribution across training, validation sets.}}
\label{tab:data_split}
\centering
\setlength{\tabcolsep}{12pt}
\renewcommand{\arraystretch}{1.5}
\begin{tabular}{lcccc}
\hline
 & \textbf{Overall, n (\%)} & \textbf{Normal NT, n (\%)} & \textbf{Cystic Hygroma, n (\%)} \\
\hline
\textbf{Total images} & 289 (100\%) & 160 (100\%) & 129 (100\%) \\
\textbf{Training set} & 217 (75.1\%) & 120 (75\%) & 97 (75.2\%)$^a$ \\
\textbf{Validation set} & 72 (24.9\%) & 40 (25\%) & 32 (24.8\%) \\
\hline
\multicolumn{4}{l}{\footnotesize
\parbox{0.9\textwidth}{$^{a}$ A final cystic hygroma training dataset of 120 images was created by randomly resampling 23 of the 97 original images to mitigate the imbalance between the two groups.
}}
\end{tabular}
\end{table*}

Images were retrospectively collected from the Institutional Picture Archiving and Communication System (PACS). The dataset comprised mid-sagittal fetal ultrasound scans collected for NT assessment as part of routine first-trimester screening exams. Cases of cystic hygroma were identified on the basis of expert clinical interpretation and sonographic features that were characterized by an enlarged, septated or multiloculated cystic mass in the posterior fetal neck region. Normal controls were characterized by NT measurements that fell within the expected range for gestational age and showed no evidence of cystic or septated abnormalities.

289 ultrasound images were included in the dataset, which included 129 cystic hygroma and 160 normal control images ({Table~\ref{tab:data_split}}). These images were included in the dataset on the basis of inclusion and exclusion criteria defined in the original study {\cite{b5}}, and images were excluded if there was poor visualization of the fetal neck region or if the scan was not acquired in the appropriate mid-sagittal plane.

The study was approved by the Ottawa Health Science Network Research Ethics Board (OHSN REB \#20210079), and the requirement for informed consent was waived on the basis of the retrospective nature of the analysis and the use of de-identified imaging data. All procedures were performed in accordance with the relevant guidelines and ethical standards.

\subsection{Data Preprocessing}
\begin{figure*}
	\centering
	\includegraphics[width=0.98\textwidth]{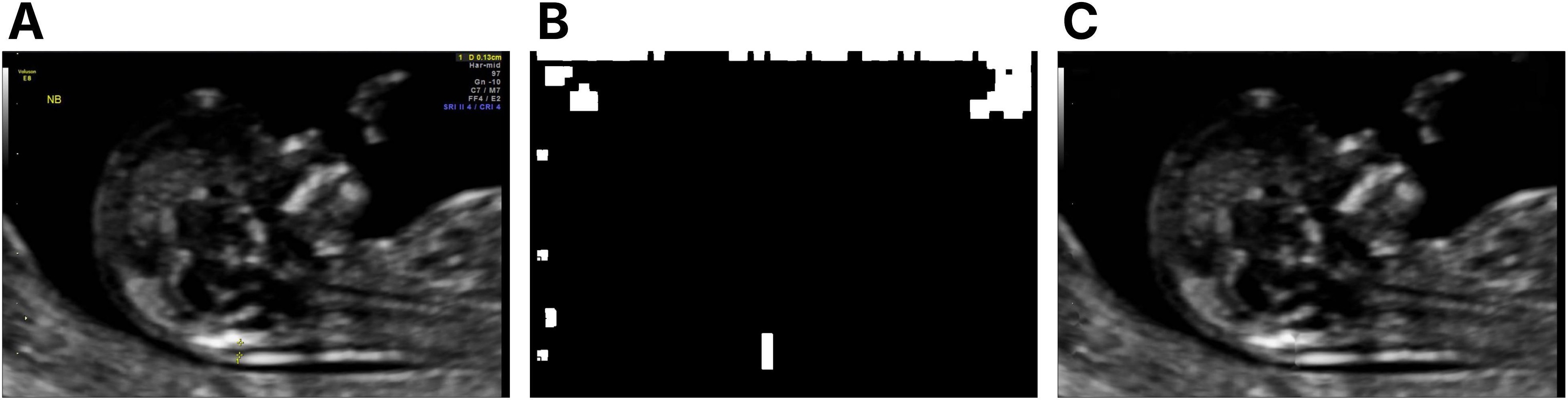}
	\caption{(A) The initial fetal ultrasound image. (B) Non-grayscale artifact locations are identified via a binary mask. (C) The image was processed using a Navier-Stokes-based {\cite{b21}} inpainting technique after artifact removal.}
	\label{FIG:annotation_removal}
\end{figure*}

All ultrasound images underwent the same preprocessing pipeline described in the original cystic hygroma study by Walker \textit{et al.} {\cite{b5}}. The raw DICOM images contained coloured annotations, including callipers, measurement text, icons ({Fig~\ref{FIG:preprocessing_example}B}), and profile traces ({Fig~\ref{FIG:annotation_removal}A}), as well as embedded patient personal health information (PHI), which were removed prior to model training and evaluation.

PHI was removed by cropping the image borders to eliminate identifying information. Coloured annotations were subsequently removed through a multi-step image processing procedure. First, images were converted from the Red-Green-Blue (RGB) colour space to the Hue-Saturation-Value (HSV) colour space. Pixels corresponding to the grayscale ultrasound image were empirically identified using threshold ranges of 0-27 for hue, 0-150 for saturation, and 0-255 for value, as shown in {Fig~\ref{FIG:preprocessing_example}}. Pixels falling outside these ranges were classified as belonging to coloured annotations.

A binary mask was then generated in which annotation pixels were assigned a value of one and ultrasound image pixels were assigned a value of zero. This mask was dilated using a $5 \times 5$ kernel to ensure that annotation contours were fully captured. The areas of removed coloured annotations were filled back in using a Navier–Stokes–based image inpainting algorithm {\cite{b21}} ({Fig~\ref{FIG:annotation_removal}}). Inpainting is modelled as a flow process, and pixel values are propagated from the boundary of the removed regions to the interior by iterative diffusion. By formulating the problem using principles from Navier–Stokes fluid dynamics, the algorithm preserves edge continuity and texture structure, yielding visually plausible reconstructions in place of the removed annotations. This approach has been widely used in medical imaging preprocessing because it maintains the local ultrasound texture statistics in the filled regions while not introducing artificial patterns that could bias model learning.

Following annotation removal, all images were converted to grayscale and intensity values were standardized to have zero mean and unit variance to improve numerical stability during network training. Finally, images were resized to a fixed spatial resolution of $224 \times 224$ pixels before being used as input to the USF-MAE model {\cite{b1}}.

\subsection{Model Training Framework}
\begin{figure*}
\centering
\includegraphics[width=0.9\textwidth]{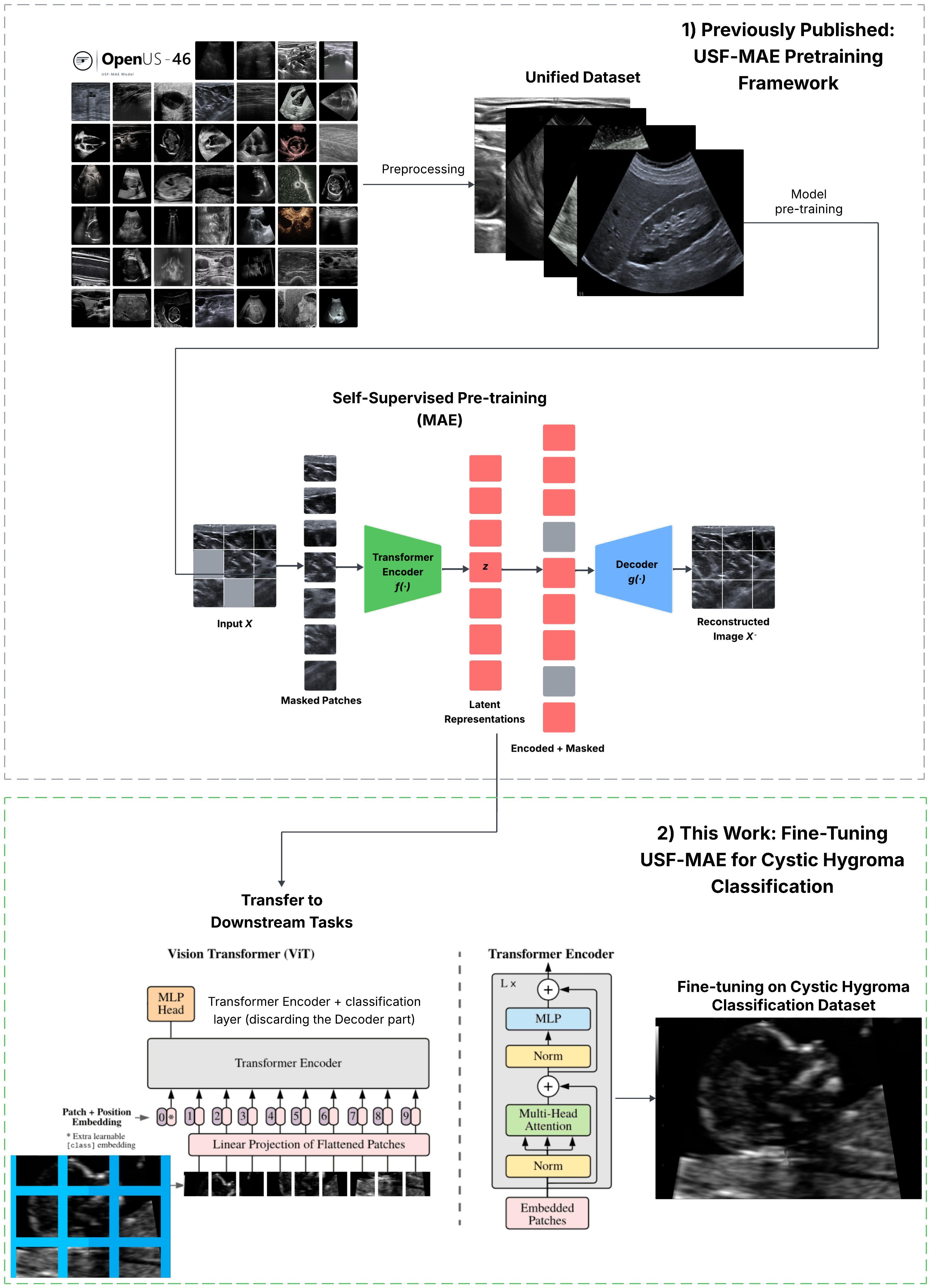}
\caption{Overview of the cystic hygroma classification workflow. The top panel shows the existing USF-MAE pretraining framework previously developed by our group {\cite{b1}}, where a large, consolidated ultrasound dataset is preprocessed and employed for SSL using MAE. Randomly sampled image patches are hidden and reconstructed during pretraining, resulting in the encoder learning ultrasound-specific feature representations. The bottom panel depicts the contribution of this study, which extends this work by adapting the pretrained encoder for cystic hygroma classification, where the MAE decoder is discarded, and a classification head is added. The model is then fine-tuned using labelled fetal neck ultrasound data for diagnostic prediction support.}
\label{FIG:model_arc}
\end{figure*}

\textbf{1) USF-MAE Architecture:}
The proposed approach is based on the Ultrasound Self-Supervised Foundation Model with Masked Autoencoding (USF-MAE) {\cite{b1}}, a transformer-based SSL framework developed by our research group for ultrasound imaging. USF-MAE follows the masked autoencoder paradigm introduced by He \textit{et al.} {\cite{b9}}, in which a ViT encoder {\cite{b26,b27}} is trained to learn meaningful representations by reconstructing masked portions of input images ({Fig~\ref{FIG:model_arc}}).

During pretraining, each ultrasound image is divided into fixed, non-overlapping $16 \times 16$ patches. A random subset corresponding to $25\%$ of the patches is masked, while the remaining visible patches are linearly embedded and processed by the ViT encoder through multi-head self-attention layers {\cite{b26,b27}}. The encoder learns to capture ultrasound-specific texture patterns and anatomical context by modelling long-range dependencies across the image. A lightweight decoder reconstructs the masked patches, and training is optimized by minimizing the mean squared error between the reconstructed and original images. Pretraining was performed on the OpenUS-46 dataset ({Fig~\ref{FIG:model_arc}}), which contains over 370,000 ultrasound images spanning 46 publicly available datasets and a wide range of anatomical regions {\cite{b1}}.

\textbf{2) Model Initialization and Adaptation for Cystic Hygroma Classification:}
For the cystic hygroma detection task, the pretrained USF-MAE encoder was used as the initialization point. The decoder, which is only required during self-supervised pretraining, was removed. A task-specific classification head consisting of a fully connected layer with two output units was attached to the encoder to enable binary classification between normal fetuses and those with cystic hygroma. A SoftMax activation was applied to convert logits into normalized class probabilities. Both the encoder and the classification head were fine-tuned jointly using supervised learning.

\textbf{3) Supervised Fine-tuning and Cross-validation Strategy:}
We trained and validated our model following exactly the same experimental protocol as the original cystic hygroma study by Walker et al. {\cite{b5}}. We used 4-fold cross-validation, dividing the dataset into four folds that were mutually exclusive. In each cross-validation fold, three were used for training and one was used for validation, until each fold had been used as a validation set exactly once. To compensate for the class imbalance between cystic hygroma and normal cases within the training folds, cystic hygroma images were randomly upsampled with replacement to match the number of normal control images in the training set, as in the original study {\cite{b5}}.

\textbf{4) Data Augmentation Strategy:}
Data augmentations were only used for the training folds, while validation folds were kept unaugmented. The augmentations used only introduced appearance variability that was anatomically plausible (so as not to introduce augmentations that may obscure diagnostic findings). In particular, training images were resized to $224 \times 224$ pixels, then rotated randomly by $0^{\circ}$ to $90^{\circ}$, randomly horizontally and vertically flipped with probability $0.5$, and randomly resized and cropped with scale $0.5$ to $2.0$. Images were then converted to tensor format and normalized with ImageNet {\cite{b18}} mean and standard deviation (SD) values. Validation images were only resized and normalized, without stochastic transformations.

\textbf{5) Optimization Strategy:}
The supervised fine-tuning was conducted with the AdamW {\cite{b22}} optimizer. We used a learning rate of $\{1 \times 10^{-3}\}$ and a weight decay value of $\{0.0001\}$. We trained the experiments for 100 epochs with a batch size of 64. A cosine learning rate schedule with linear warm-up over the first 10\% of epochs was used. The model checkpoint with the highest validation accuracy within each fold was kept for evaluation.

\textbf{6) Computational Environment and Implementation Details:}
The model was implemented with the PyTorch DL framework. All experiments were run on a SLURM-managed high-performance research computing cluster with NVIDIA L40S GPUs. Training one cross-validation fold took around 15 minutes over 100 epochs. Checkpoints, training logs, and configuration files were saved for each fold. The pretrained weights of the USF-MAE model, along with the OpenUS-46 ultrasound image dataset, are publicly accessible through our repository (\href{https://github.com/Yusufii9/USF-MAE}{\textcolor{blue}{GitHub Repository}}: https://github.com/Yusufii9/USF-MAE).

\subsection{Baseline Model: DenseNet-169}
To provide a CNN baseline for comparison, the proposed USF-MAE framework was evaluated against the DenseNet-169 architecture originally developed and validated for cystic hygroma classification by Walker \textit{et al.} {\cite{b5}}. DenseNet {\cite{b23}} is a family of CNNs characterized by dense connectivity, in which each layer receives feature maps from all preceding layers through direct connections {\cite{b18}}. This design promotes feature reuse, improves gradient flow, and enables efficient parameter utilization, which has been shown to be advantageous for medical image analysis tasks, including ultrasound imaging.

In the original study, a PyTorch implementation of DenseNet-169 was employed. The input layer was modified to accept grayscale ultrasound images with a spatial resolution of $256 \times 256$ pixels, and the final classification layer was replaced with a fully-connected layer producing two output classes: normal and cystic hygroma. DenseNet-169 was selected over alternative architectures such as ResNet due to its favourable balance between classification performance and computational efficiency, as previously demonstrated by Gao \textit{et al.} {\cite{b23}}.

The DenseNet-169 model was trained from scratch with random weight initialization, without using any pretrained weights from natural image datasets such as ImageNet {\cite{b18}}. This design choice was motivated by the substantial domain mismatch between natural images and fetal ultrasound, which limits the effectiveness of conventional transfer learning in this context. Model training was performed using the cross-entropy loss function and the Adam optimizer {\cite{b24}}, with optimizer parameters set to $\beta_1 = 0.9$, $\beta_2 = 0.999$, and $\epsilon = 1 \times 10^{-8}$. A batch size of 64 was used throughout training.

The network was trained for 1000 epochs using a step-based learning rate schedule. The initial learning rate was set to $1 \times 10^{-2}$ and decayed by a factor of 0.72 every 100 epochs using a learning rate step scheduler. Data augmentation was applied dynamically during training to improve generalization and reduce overfitting. Augmentation operations included random horizontal flipping with 50\% probability, random rotations in the range of $[-15^{\circ}, 15^{\circ}]$, random translations (up to $\pm10\%$ horizontally and $\pm30\%$ vertically of the image size), and random shearing in the range of $[-0.2^{\circ}, 0.2^{\circ}]$.

To address class imbalance in the training data, cystic hygroma images were randomly upsampled with replacement to match the number of normal control images. The DenseNet-169 model was trained and evaluated using the same 4-fold cross-validation strategy and preprocessing pipeline as described before.

No modifications were made to the DenseNet-169 architecture, training procedure, or hyperparameters beyond those reported by Walker \textit{et al.} {\cite{b5}}. This ensured that performance differences observed between DenseNet-169 and the proposed USF-MAE model reflect the impact of ultrasound-specific self-supervised pretraining rather than differences in experimental configuration.

\subsection{Performance Metrics}
For the quantitative assessment of our proposed USF-MAE framework and the DenseNet-169 baseline, we use the same metrics reported in Walker \textit{et al.}'s original cystic hygroma study {\cite{b5}}. The metrics of classification accuracy, sensitivity, specificity, and area under the receiver operating characteristic curve (ROC-AUC) are used. The combination of these metrics describe both the overall classification performance and model's ability to correctly classify cases of cystic hygroma, while minimizing the rates of false-negative and false-positive predictions.

All metrics were calculated on the validation set for each fold in the 4-fold cross-validation. Performance is reported at the epoch corresponding to the highest validation accuracy. Final reported performance is averaged across the four folds.

\textbf{1) Accuracy:}
Accuracy represents the overall proportion of correct predictions made by the model over both the positive and negative classes. It is defined as:
\begin{equation}
\text{Accuracy} = \frac{TP + TN}{TP + TN + FP + FN}
\end{equation}
Where $TP$, $TN$, $FP$, and $FN$ refer to the number of true positives, true negatives, false positives, and false negatives, respectively. Accuracy is a single scalar measure of general model performance. However, its value can be biased by class imbalance, so it is considered in the context of sensitivity and specificity.

\textbf{2) Sensitivity (Recall):}
Sensitivity is a measure of the model's performance at identifying fetuses with cystic hygroma correctly:
\begin{equation}
\text{Sensitivity} = \frac{TP}{TP + FN}
\end{equation}
High sensitivity is particularly important in the context of screening prenatal ultrasound images for cystic hygroma, as false negative cases may lead to a delay in further genetic testing, counselling, or pregnancy management.

\textbf{3) Specificity:}
Specificity is the proportion of normal cases correctly classified by the model:
\begin{equation}
\text{Specificity} = \frac{TN}{TN + FP}
\end{equation}
High specificity is important to limit the number of false positive findings, which may otherwise lead to additional follow-up imaging studies, invasive diagnostic procedures, or parental anxiety.

\begin{table*}
\caption{\textbf{4-fold cross-validation results for DenseNet-169 (baseline) vs. USF-MAE.}}
\centering
\begin{tabular}{lccccc|ccccc}
\hline
 & \multicolumn{5}{c}{DenseNet-169 {\cite{b5}}} & \multicolumn{5}{c}{\textbf{USF-MAE (new)}} \\

\textbf{Metric} & \textbf{Fold 1} & \textbf{Fold 2} & \textbf{Fold 3} & \textbf{Fold 4} & \textbf{Mean$\pm$SD} & \textbf{Fold 1} & \textbf{Fold 2} & \textbf{Fold 3} & \textbf{Fold 4} & \textbf{Mean$\pm$SD} \\
\hline
\textbf{Accuracy} & 0.89 & 0.93 & 0.96 & 0.94 & 0.93$\pm$0.03 & \textbf{0.96} & \textbf{0.93} & \textbf{0.97} & \textbf{0.97} & \textbf{0.96$\pm$0.02} \\
\textbf{Sensitivity} & 0.82 & \textbf{0.91} & 1.00 & 0.97 & 0.92$\pm$0.07 & \textbf{0.91} & 0.88 & \textbf{1.00} & \textbf{0.97} & \textbf{0.94$\pm$0.06} \\
\textbf{Specificity} & 0.95 & 0.95 & 0.93 & 0.93 & 0.94$\pm$0.01 & \textbf{1.00} & \textbf{0.98} & \textbf{0.95} & \textbf{0.98} & \textbf{0.98$\pm$0.02} \\
\textbf{AUC} & 0.91 & 0.93 & 0.98 & 0.95 & 0.94$\pm$0.03 & \textbf{0.97} & \textbf{0.95} & \textbf{1.00} & \textbf{1.00} & \textbf{0.98$\pm$0.02} \\
\hline
\end{tabular}
\label{tab:comparison_results}
\end{table*}

\textbf{4) Area Under the Receiver Operating Characteristic Curve (ROC-AUC):}
ROC-AUC is a threshold-independent evaluation metric that measures the trade-off between sensitivity and false positive rate for all possible classification thresholds. It is calculated by plotting the true positive rate "TPR" (sensitivity) against the false positive rate "FPR" ($1 - \text{specificity}$). The area under the ROC curve (AUC) ranges from 0.5, which indicates a model with no discriminative ability, to 1.0, for a perfect model. AUC can thus be interpreted as a single scalar value that represents the model's ability to produce consistently higher confidence scores for cystic hygroma cases relative to normal controls.

\subsection{Statistical Analysis}
A paired nonparametric Wilcoxon signed-rank test {\cite{b34}} was applied to the proposed USF-MAE performance measurements and the DenseNet-169 baseline measurements in order to verify whether the performance differences between the two approaches were statistically significant. Paired performance measurements across the cross-validation folds were considered, treating cross-validation folds as matched samples. This choice of the test was motivated by a small sample size and the desire to make no distribution assumptions. A two-sided test was used with a significance level set to $p < 0.05$.

\subsection{Explainable AI Framework}
To assess the interpretability of the proposed model and verify that predictions were driven by clinically relevant image regions, Score-CAM was employed as an explainable AI technique {\cite{b25}}. Score-CAM generates class activation maps by measuring the contribution of individual feature maps to the model's output score, without relying on gradient information.

Score-CAM heatmaps were generated for correctly classified examples from both normal and cystic hygroma classes. These visualizations were overlaid on the original ultrasound images to qualitatively assess whether the model focused on the fetal neck and NT region, which is the anatomically relevant area for cystic hygroma diagnosis. The explainability analysis was used solely for qualitative assessment and did not influence model training or evaluation.

\section{Results}
We assessed the performance of the proposed USF-MAE and compared it to the DenseNet-169 baseline in a 4-fold cross-validation setting. Table~\ref{tab:comparison_results} reports the classification accuracy, sensitivity, specificity, and AUC for each fold and mean ($\pm$ SD) across folds.

Across all evaluation metrics, the USF-MAE model outperformed the DenseNet-169 baseline consistently. In terms of classification accuracy, USF-MAE achieved a mean of $0.96 \pm 0.02$ compared to $0.93 \pm 0.03$ for DenseNet-169. Improvements in classification accuracy were observed in three of the four folds, with USF-MAE achieving accuracies of 0.96, 0.97, and 0.97 in Folds 1, 3, and 4, respectively.

Sensitivity, or the ability to correctly identify cases of cystic hygroma, was also higher in the USF-MAE model. The USF-MAE model achieved a mean sensitivity of $0.94 \pm 0.06$ compared to $0.92 \pm 0.07$ for DenseNet-169. Both models achieved perfect sensitivity (1.00) in Fold 3, but USF-MAE demonstrated more stable sensitivity across folds, particularly in Fold 1 and Fold 4.

The largest improvements were observed in specificity, with the USF-MAE model achieving a mean of $0.98 \pm 0.02$ compared to $0.94 \pm 0.01$ for DenseNet-169. Notably, USF-MAE achieved perfect specificity (1.00) in Fold 1 and maintained specificity values of 0.95 or higher across all folds, indicating a lower false-positive rate compared to the baseline model.

The ROC-AUC analysis further demonstrated the superior discriminative performance of the USF-MAE model. The new proposed model achieved a mean AUC of $0.98 \pm 0.02$, surpassing the DenseNet-169 mean of $0.94 \pm 0.03$. USF-MAE achieved an AUC of 1.00 in two of the four folds, indicating near-perfect separation between normal and cystic hygroma cases in those validation splits.

A paired Wilcoxon signed-rank test {\cite{b34}} comparing per-fold performance between USF-MAE and DenseNet-169 demonstrated a statistically significant improvement in favour of the proposed model. Across all evaluated metrics, USF-MAE consistently achieved higher scores than the baseline. The Wilcoxon test yielded a test statistic of $W = 6.0$ with a corresponding two-sided $p$-value of $p = 0.0057$ ($p < 0.05$), indicating that the observed performance gains were unlikely to have occurred by chance.

In summary, across all evaluation metrics, the USF-MAE not only achieved higher mean performance than DenseNet-169 but also exhibited lower fold-to-fold variability, indicating improved robustness and generalization when fine-tuned on the limited cystic hygroma dataset. These results also indicate that the improvements achieved are statistically significant at the 5\% level.

\begin{figure*}
	\centering
	\includegraphics[width=0.98\textwidth]{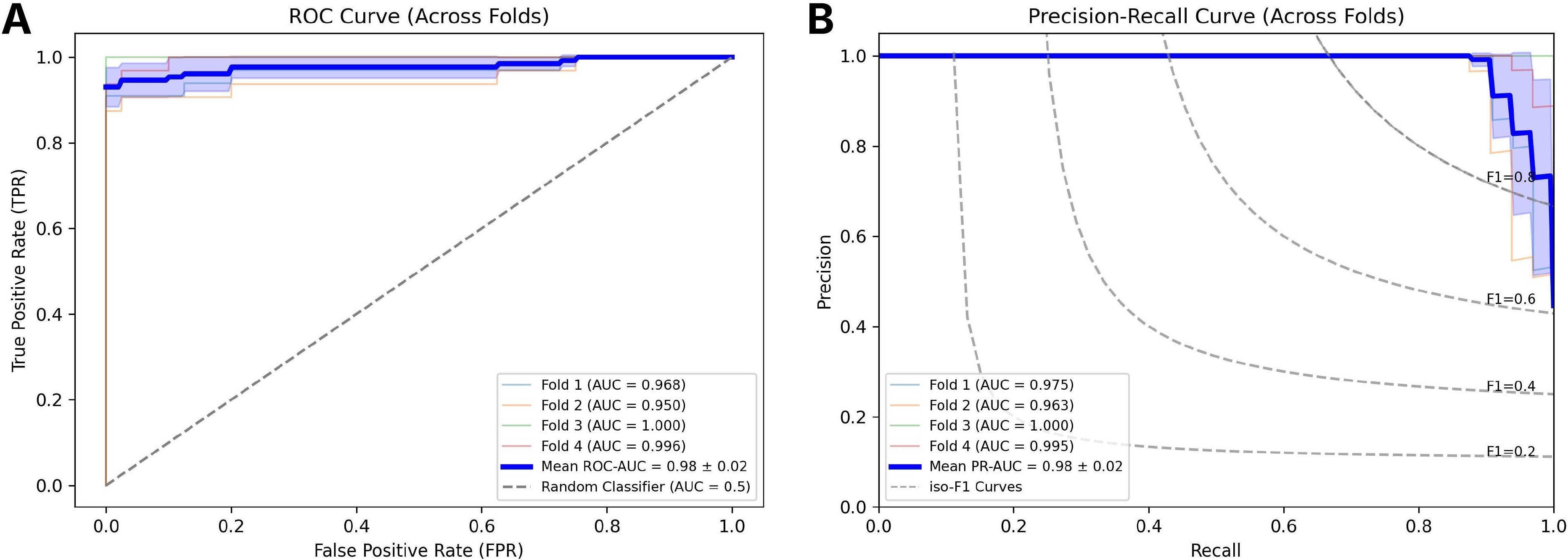}
	\caption{\textbf{ROC and precision-recall (PR) curves across cross-validation folds.}
    (A) ROC curves for each of the four cross-validation folds, along with the mean ROC curve (solid blue line) and its variability. The dashed diagonal line represents the performance of a random classifier. The proposed model achieves a high mean ROC-AUC across folds, indicating strong discriminative performance.
    (B) Precision-recall curves across folds with the mean PR curve shown in blue. Iso-F1 score contours are overlaid to illustrate the trade-off between precision and recall. The consistently high PR-AUC indicates robust performance under class-imbalanced conditions.}
	\label{FIG:model_curves}
\end{figure*}

\section{Discussion}
We presented a proof-of-concept application of ultrasound-specific self-supervised fine-tuning for the automatic detection of cystic hygroma from first-trimester fetal ultrasound images. We leveraged the USF-MAE pretrained encoder to fine-tune on the same curated dataset as Walker \textit{et al.} {\cite{b5}} and found that ultrasound-specific self-supervised pretraining consistently improved diagnostic performance over a CNN trained longer but from scratch. Across all metrics, USF-MAE outperformed the DenseNet-169 baseline, providing evidence to support the hypothesis that modality-specific representation learning can provide benefits for robustness and generalization in data-limited prenatal imaging tasks.

\subsection{Improved Diagnostic Performance for Cystic Hygroma Detection}
USF-MAE outperformed DenseNet-169 in all four cross-validation folds for each performance metric: accuracy, sensitivity, specificity, and AUC ({Table~\ref{tab:comparison_results}}). In particular, USF-MAE achieved a mean ROC-AUC of $0.98 \pm 0.02$, significantly higher than the DenseNet-169 baseline of $0.94 \pm 0.03$. This suggests strong discriminative power between normal fetuses and those with cystic hygroma for the proposed model.

The ROC curves in {Fig~\ref{FIG:model_curves}A} show a consistently high true positive rate at low false positive rates across all validation folds and low variability between folds. This suggests that the proposed model generalizes well across different cross-validation folds and is able to reliably learn a decision boundary for this task despite the modest dataset size. In parallel, the PR curves in {Fig~\ref{FIG:model_curves}B} show consistently high performance across recall thresholds, with a mean PR-AUC of $0.98 \pm 0.02$. This further emphasizes the proposed model's ability to achieve high precision even at high sensitivity.

In the clinical context, these improvements in sensitivity and specificity are especially meaningful as cystic hygroma represents a high-risk prenatal finding with established associations to chromosomal abnormalities {\cite{b29,b31,b32}}, structural malformations {\cite{b28,b30,b31}}, and poor pregnancy outcomes {\cite{b29,b30}}. Improved sensitivity minimizes the risk of false negatives, while improved specificity reduces the risk of unnecessary downstream testing and parental anxiety.

\subsection{Robustness Across Cross-Validation Folds}
A notable observation in this study is the lower variability in performance metrics across the different cross-validation folds for USF-MAE compared to DenseNet-169. Although both models achieved high performance in some folds, USF-MAE produced more consistent and reliable results, especially in specificity and ROC-AUC. This is a strong indication that the pretrained encoder provides a stable initialization and regularization signal to the learning process, enabling reliable performance even in the low data regime.

\begin{figure*}
	\centering
	\includegraphics[width=0.98\textwidth]{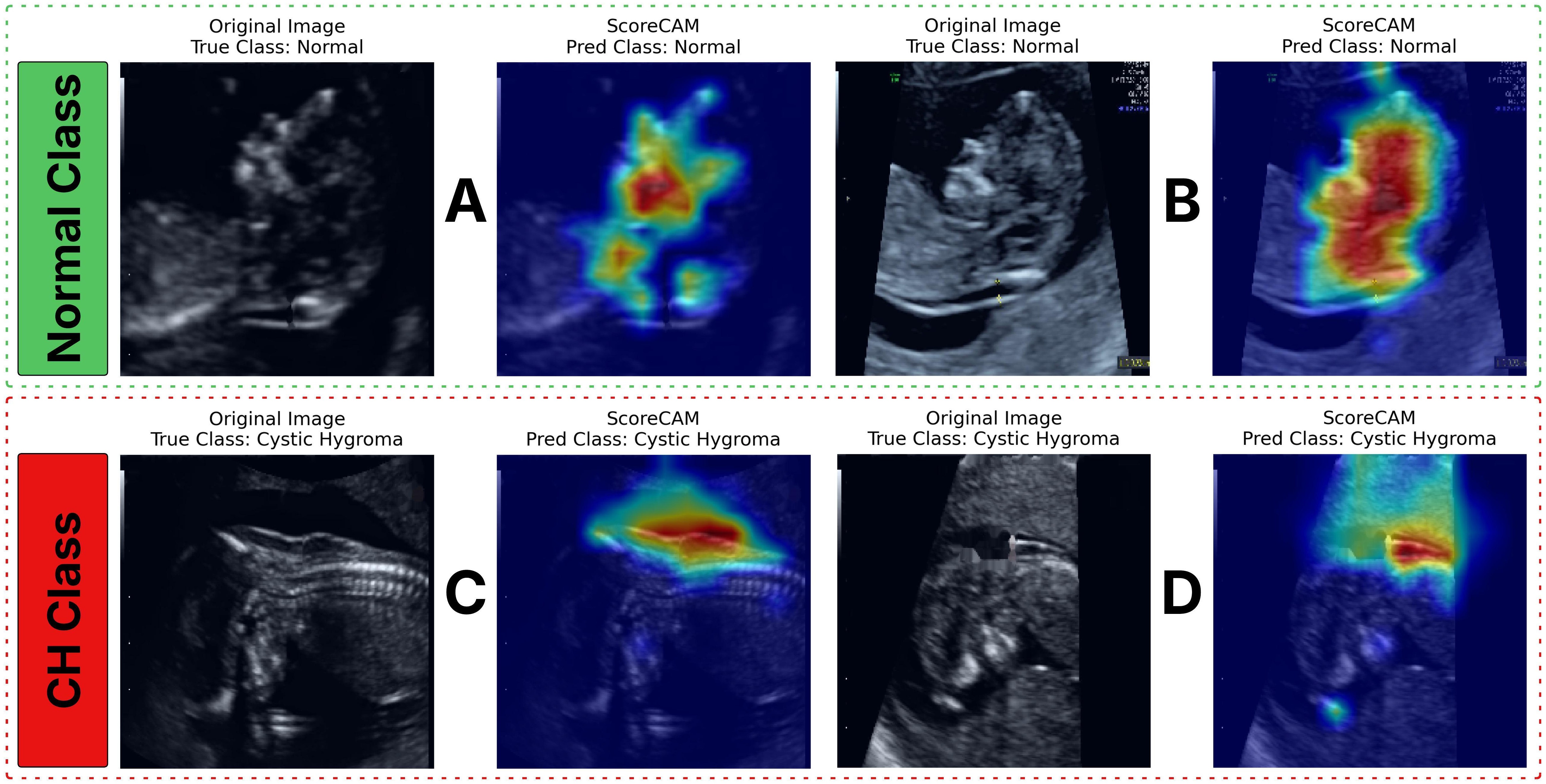}
	\caption{\textbf{Score-CAM visual explanations for model predictions.} Representative ultrasound images and corresponding Score-CAM {\cite{b25}} activation maps illustrating regions contributing to the model's predictions.
    (A-B) Examples from the normal class, where salient activations are distributed over anatomically relevant fetal regions without focal pathological emphasis.
    (C-D) Examples from the cystic hygroma (CH) class, where Score-CAM highlights localized regions corresponding to fluid-filled cystic structures in the fetal neck.
    These visualizations demonstrate that the model focuses on clinically meaningful regions when making classification decisions, supporting interpretability and trustworthiness.}
	\label{FIG:XAI_figure}
\end{figure*}

This improved robustness can be attributed to the self-supervised pretraining strategy on a large, heterogeneous corpus of ultrasound volumes. By learning ultrasound-specific texture, speckle patterns, and anatomical context in the pretraining step, the model leverages information from a much larger source of data compared to the downstream dataset alone. This pretraining signal during the crucial initial layers of feature learning provides an effective regularizer and enables robust downstream fine-tuning. In contrast, training DenseNet-169 from random initialization requires the limited cystic hygroma dataset to support all stages of feature learning.

\subsection{Explainability and Anatomical Plausibility}
Explainability is critical for the clinical adoption of AI systems in prenatal imaging. To assess whether the USF-MAE model relied on clinically meaningful image regions, Score-CAM visualizations {\cite{b25}} were generated for representative normal and cystic hygroma cases ({Fig~\ref{FIG:XAI_figure}}). 

For normal cases ({Fig~\ref{FIG:XAI_figure}A,B}), the model's attention was distributed over the expected fetal head and neck region without focusing on spurious background structures. In cystic hygroma cases ({Fig~\ref{FIG:XAI_figure}C,D}), Score-CAM heatmaps consistently highlighted the posterior fetal neck and nuchal region, corresponding to the enlarged, septated fluid collections that define cystic hygroma. These visualizations align closely with clinical reasoning and mirror the qualitative findings reported in the original DenseNet-based study {\cite{b5}}.

Importantly, the use of a transformer-based architecture did not compromise explainability. Instead, the model's attention maps suggest that USF-MAE learned anatomically grounded features that are directly relevant to the diagnostic task, rather than relying on imaging artifacts or global intensity patterns.

\subsection{Implications of Ultrasound-Specific Self-Supervised Pretraining}
The performance gains observed in this study further support the value of ultrasound-specific self-supervised learning for prenatal imaging applications. Unlike traditional transfer learning approaches that rely on natural image datasets such as ImageNet, USF-MAE is pretrained exclusively on ultrasound data, enabling it to capture modality-specific characteristics, such as speckle noise, low-contrast boundaries, and operator-dependent variability.

In the context of cystic hygroma detection, where labelled datasets are inherently small due to the condition's rarity, leveraging large amounts of unlabelled ultrasound data is particularly advantageous. The improved performance of USF-MAE relative to an optimized CNN baseline trained from scratch highlights the limitations of purely supervised learning in this setting and underscores the importance of foundation models tailored to medical imaging modalities.

\subsection{Data Efficiency and Practical Advantages of Self-Supervised Pretraining}
In addition to the performance gains reported earlier in this section, another significant benefit of the USF-MAE framework we proposed is the use of large-scale unlabeled ultrasound data for pretraining. Unlike a fully supervised learning paradigm, self-supervised pretraining is not contingent on manual annotations, which are expensive and labour-intensive in healthcare contexts. Manual labelling of ultrasound images by experts in prenatal screening requires clinical domain expertise, careful curation, and time from trained personnel such as sonographers or maternal–fetal medicine specialists. The latter factors pose nontrivial bottlenecks to scaling supervised learning, especially for rare conditions like cystic hygroma. However, by pretraining on unlabelled images, USF-MAE circumvents this issue and enables the exploitation of routine clinical data that would otherwise have been unavailable for model pretraining. Large tertiary care centers alone are amassing large volumes of ultrasound scans routinely as part of their clinical workflow, the majority of which are never annotated beyond routine reporting and hence remain essentially unexploited for model development. The opportunity to leverage such data for SSL of representations opens up a practical, cost-effective route to train more robust models without incurring the heavy annotation cost.

In terms of clinical workflow, this framework also has the potential to enable more flexible, faster iterations and deployments of such AI tools. As more ultrasound data become available, the foundation model could continue to be enriched with additional unlabelled examples to improve representations, without altering the clinical workflow. Upon task-specific fine-tuning for target conditions such as cystic hygroma, a relatively small amount of labelled data suffices, which makes this learning strategy very attractive for rare or high-risk prenatal findings. In the long run, this data-efficient strategy can fast-track the development of AI-enabled screening tools to assist ultrasound operators in their day-to-day work. By serving as a reliable source of decision support, self-supervised ultrasound foundation models like USF-MAE can help reduce operator dependence, improve consistency in early anomaly detection, and make prenatal screening programs more efficient overall.

\subsection{Limitations}
Several limitations should be considered when interpreting the findings of this study. First, the dataset was derived from a single tertiary care institution and consisted of images acquired using a limited set of ultrasound systems. Differences in image appearance across vendors, acquisition protocols, and clinical sites may affect generalizability. External validation across multiple institutional datasets will be necessary to fully assess the robustness of the proposed framework.

Second, while cross-validation was used to mitigate overfitting, the overall dataset remains modest in size. Although self-supervised pretraining reduced sensitivity to data scarcity, further performance improvements may be achievable with larger and more diverse training cohorts. Finally, this study focused on still-frame images from mid-sagittal views. Incorporating temporal cine loops or three-dimensional ultrasound data may provide additional contextual information, further enhancing diagnostic accuracy.

\subsection{Future Work}
Future research should explore the generalizability of the USF-MAE framework to external datasets acquired across different institutions and ultrasound vendors. Extending the approach to additional first-trimester screening tasks, such as automated nuchal translucency measurement or detection of other early structural anomalies, represents a natural next step. Incorporating uncertainty estimation into model predictions may further improve clinical utility by identifying cases that warrant closer manual review. Ultimately, prospective evaluation in a real-world clinical workflow will be essential to determine the practical impact of this approach on prenatal screening and decision-making.

\section{Conclusion}
In our experiments, we have shown that fine-tuning an ultrasound-specific self-supervised foundation model yields accurate and robust detection of cystic hygroma in first-trimester fetal ultrasound images. By fine-tuning the USF-MAE pretrained encoder (\href{https://github.com/Yusufii9/USF-MAE}{\textcolor{blue}{GitHub Repository}}: https://github.com/Yusufii9/USF-MAE) on the same dataset and experimental protocol of a previous convolutional baseline, the proposed method has consistently outperformed DenseNet-169 across all performance metrics (accuracy, sensitivity, specificity and ROC-AUC). These results support the benefits of ultrasound-specific self-supervised pretraining, especially in data-constrained clinical scenarios where expert annotations can be expensive and limited. The better performance and robustness of USF-MAE, as well as the anatomically meaningful explainability provided by Score-CAM, may help establish its trustworthiness as a potential decision-support tool for early prenatal anomaly screening. In summary, this study has demonstrated the potential of self-supervised ultrasound foundation models as a scalable, data-efficient approach for AI-assisted obstetric imaging and the early detection of high-risk fetal conditions.



\end{document}